\documentclass[sn-mathphys,Numbered]{sn-jnl}

\usepackage{graphicx}%
\usepackage{multirow}%
\usepackage{amsmath,amssymb,amsfonts}%
\usepackage{amsthm}%
\usepackage{mathrsfs}%
\usepackage[title]{appendix}%
\usepackage{xcolor}%
\usepackage{textcomp}%
\usepackage{manyfoot}%
\usepackage{booktabs}%
\usepackage{algorithm}%
\usepackage{algorithmicx}%
\usepackage{algpseudocode}%
\usepackage{listings}%

\usepackage{caption}

\usepackage{tcolorbox}

\usepackage{wasysym}
\usepackage{array}

\begin{document}

\title[Article Title]{Classification of Human- and AI-Generated Texts:\\Investigating Features for ChatGPT}


\author{\fnm{Lorenz} \sur{Mindner}}

\author{\fnm{Tim} \sur{Schlippe}}\email{tim.schlippe@iu.org}

\author{\fnm{Kristina} \sur{Schaaff}}\email{kristina.schaaff@iu.org}

\affil{\orgdiv{IU International University of Applied Sciences}, 
\orgaddress{
\country{Germany}}}




\abstract{Recently, generative AIs like ChatGPT have become available to the wide public. These tools can for instance be used by students to generate essays or whole theses. But how does a teacher know whether a text is written by a student or an AI? In our work, we explore traditional and new features to (1) detect text generated by AI from scratch and (2) text rephrased by AI. Since we found that classification is more difficult when the AI has been instructed to create the text in a way that a human would not recognize that it was generated by an AI, we also investigate this more \textit{advanced} case. For our experiments, we produced a new text corpus covering 10 school topics. Our best systems to classify \textit{basic} and \textit{advanced} \textit{human-generated}/\textit{AI-generated} texts have F1-scores of over~96\%. Our best systems for classifying \textit{basic} and \textit{advanced} \textit{human-generated}/\textit{AI-rephrased} texts have F1-scores of more than~78\%. The systems use a combination of perplexity, semantic, list lookup, error-based, readability, AI feedback, and text vector features. Our results show that the new features substantially help to improve the performance of many classifiers. Our best \textit{basic} text rephrasing detection system even outperforms GPTZero by 183.8\% relative in F1-score.

}




\maketitle


\section{Introduction}\label{sec1}

In recent years, chatbots have become a popular tool in everyday life \cite{pelau2021makes}. These systems are capable of imitating human-like conversations with users~\cite{adiwardana2020towards}, can provide assistance~\cite{dibitonto2018chatbot}, information~\cite{arteaga2019design}, and emotional support~\cite{falala2019owlie}. OpenAI's ChatGPT has become one of the most commonly utilized chatbots out of all of them. The fact, that ChatGPT was able to reach over one million users in only five days~\cite{taecharungroj2023can} undermines this statement. The users of ChatGPT range from children seeking assistance with their homework, and individuals searching for medical advice, to users who use it as a daily source of companionship. 
The more systems like ChatGPT make their way into our everyday lives, the more important it becomes to differentiate between human- and artificial intelligence (AI)-generated content. Although both can communicate information, an important difference lies in the intent of the text. \textit{Human-generated} content is created with the specific intention of communicating something, while \textit{AI-generated} content is created by algorithms designed to generate text that sounds like it was written by a human. \textit{AI-generated} text may contain repetitive or formulaic phrases or patterns, while \textit{human-generated} text is more likely to be original and creative. Moreover, texts generated by large language models (LLMs) often sound reliable even though they are based on word probabilities instead of facts. The better the algorithms of generative AI become, the more difficult it is to detect \textit{AI-generated} content properly. This poses serious problems in many areas, including plagiarism, the generation of fake news, and spamming. Therefore, there is a strong need for tools that can differentiate between these two kinds of texts.


For this reason, in our current study, we want to gain insights into the differences between human language use and \textit{AI-generated} text, and how these differences can be leveraged to improve the accuracy of the detection of \textit{AI-generated} text. Furthermore, this research will provide a valuable benchmark for future \textit{AI-generated} text classification studies. To the best of our knowledge, we are the first to evaluate features such as the degree of objectivity of a text, list lookup features like the repetitions of the title in the text, or error-based features such as the number of grammatical errors. We collected a new corpus of nearly 500 articles covering 10 topics---the \textit{Human-AI-Generated Text Corpus}. To contribute to the improvement of low-resource languages, we share the corpus with the research community\footnote{\url{https://github.com/LorenzM97/human-AI-generatedTextCorpus}}. 
We decided to use ChatGPT for our research as this is currently the most widely used tool to generate texts. Since it has been trained on extensive data sets and has a huge number of parameters it is currently the best-performing system that is publicly available.  


\section{Related Work}\label{related work}

In this chapter, we will describe the related work regarding ChatGPT and the classification of \textit{human-} and \textit{AI-generated} texts.

\subsection{ChatGPT}

ChatGPT is an advanced chatbot developed by OpenAI that leverages natural language processing to generate text in response to user prompts, making it a multi-functional tool across various domains. ChatGPT has been successfully applied in domains such as education \cite{baidoo2023education}, medicine \cite{jeblick2022chatgpt}, or language translation \cite{jiao2023chatgpt}. 

As the name indicates, ChatGPT is built on the Generative Pretrained Transformers (GPT) language model and was fine-tuned using reinforcement learning with human feedback, enabling it to grasp the meaning and intention behind user queries and provide relevant and helpful responses. A large dataset of text data was incorporated into the training of ChatGPT to ensure safety and accuracy in the text generated. Although the exact amount of training data for ChatGPT has not been published, the previous GPT-3 model had 175 billion parameters and was trained with 499 billion crawled text tokens, which is substantially larger than other language models \cite{GPT3:2020} like Bidirectional Encoder Representations from Transformers (BERT)~\cite{devlin2018bert}, Robustly Optimized BERT Pretraining Approach (RoBERTa)~\cite{liu2019roberta}, or Text-to-Text Transfer Transformer (T5) \cite{roberts2019t5}. By learning the nuances of human language from this extensive dataset, ChatGPT is able to generate text that is hard to distinguish from text written by humans \cite{mitrovic2023chatgpt}.


\subsection{Classification of Human- and AI-Generated Texts}

The more ChatGPT is used in various contexts and its abilities improve, the more it becomes important to be able to identify \textit{AI-generated} and \textit{human-generated} texts. As the quality of \textit{AI-generated} texts improves, human capabilities to detect generated texts can already be outperformed by machines \cite{soni2023comparing}. Numerous tools like GPTZero\footnote{\url{https://gptzero.me}}, AI Content Detector\footnote{\url{https://writer.com/ai-content-detector}}, or GPT-2 Output Detector\footnote{\url{https://openai-openai-detector.hf.space}} exist which aim to find out if a text has been \textit{AI-generated}. These tools are based on analyzing text patterns. For instance, GPTZero---which is amongst the most popular AI-detection tools---uses perplexity and burstiness to identify \textit{AI-generated} texts. However, these tools still have limitations in terms of the detection accuracy~\cite{openai2023new}. 

In recent studies, approaches like XGBoost~\cite{Shijaku:2023}, decision trees~\cite{zaitsu2023distinguishing}, or transformer-based models~\cite{mitrovic2023chatgpt,guo2023close} have been evaluated to detect \textit{AI-generated} texts. 
\cite{mitrovic2023chatgpt} discussed several characteristics of \textit{AI-generated} texts from customer reviews and built a transformer-based classifier that was able to differentiate \textit{AI-generated} text from \textit{human-generated} text with an accuracy of 79\%. In an analysis based on decision trees, \cite{zaitsu2023distinguishing} were able to achieve an overall accuracy of 100\% combining several stylometric features (bigrams, positioning of commas, and the rate of function words) in differentiating \textit{AI-generated} from \textit{human-generated} texts. However, these analyses were limited to the Japanese language which has very different characteristics from the English language. \cite{Shijaku:2023} addressed the issue of generated essays. The proposed model was based on XGBoost and was able to achieve an accuracy of 98\% using features generated by TF-IDF and a set of handcrafted features. In a comparison of text summarizations, \cite{soni2023comparing} were able to achieve an accuracy of 90\% for the classification of \textit{AI-generated} vs. \textit{human-generated} summaries using DistilBERT. 

One of the major downsides of the previously described studies is that they have only been tested on texts which have been generated with \textit{basic} prompts asking ChatGPT to simply generate or rephrase a text. To the best of our knowledge, more \textit{advanced} prompts which ask ChatGPT to generate a text in a certain way (e.g., in a way a human does not notice) have not been included in the evaluation. 



\section{Our Human-AI-Generated Text Corpus}\label{sec:experimental setup}

To derive our statistics from the features and train models for the classification of \textit{human-} and \textit{AI-generated} text, we leverage Wikipedia articles to generate an English text corpus---our \textit{Human-AI-Generated Text Corpus}. Since the focus of our work is to recognize whether texts in an educational environment were written by humans or AI, we built a data corpus that covers various topics from this environment. For this purpose, we defined the following 10 text categories: \textit{biology}, \textit{chemistry}, \textit{geography}, \textit{history}, \textit{IT}, \textit{music}, \textit{politics}, \textit{religion}, \textit{sports}, and \textit{visual arts}. For every text category, we selected 10 topics. 
Moreover, we used different ways to generate text. Firstly, we generated texts in a \textit{basic} way without additional instructions. Secondly, texts were generated in an \textit{advanced} way using additional instructions for the generation. We also evaluated the rephrasing of texts in the same way. The following sections describe how we built our text corpus.

\subsection{Basic AI-Generated Texts}

When identifying \textit{AI-generated} text, we wanted to recognize (1) text that was generated entirely by an AI (\textit{AI-generated}), and (2) text that was rephrased by an AI based on an existing text (\textit{AI-rephrased}). 

\begin{figure}[ht]
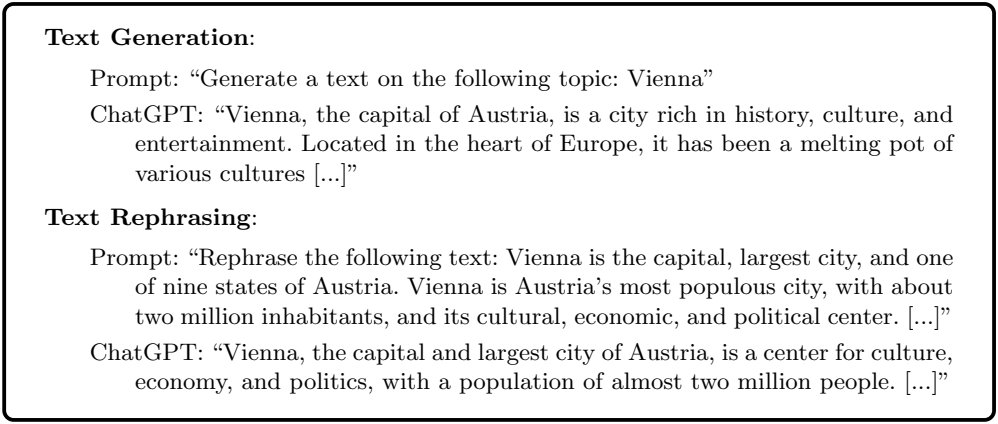

\small
  \begin{tcolorbox}[colback=white,colframe=black]
    \textbf{Text Generation}:
\vspace{0.15cm} 

\setlength{\leftskip}{0.6cm}
\hangafter=1
\hangindent=0.6cm
\noindent
Prompt: ``Generate a text on the following topic: Vienna''\vspace{0.1cm}

\hangafter=1
\hangindent=0.6cm
\noindent
ChatGPT: ``Vienna, the capital of Austria, is a city rich in history, culture, and entertainment. Located in the heart of Europe, it has been a melting pot of various cultures [...]''

\setlength{\leftskip}{0cm}
\vspace{0.2cm} 
\textbf{Text Rephrasing}:  
\vspace{0.15cm}    

\setlength{\leftskip}{0.6cm}
\hangafter=1
\hangindent=0.6cm
\noindent
Prompt: ``Rephrase the following text: Vienna is the capital, largest city, and one of nine states of Austria. Vienna is Austria's most populous city, with about two million inhabitants, and its cultural, economic, and political center. [...]''\vspace{0.1cm} 

\hangafter=1
\hangindent=0.6cm
\noindent
ChatGPT: ``Vienna, the capital and largest city of Austria, is a center for culture, economy, and politics, with a population of almost two million people. [...]''   
    
  \end{tcolorbox}
  \caption{Prompt and ChatGPT's Response: Basic Text Generation \& Rephrasing.}
  \label{fig:basicGeneration}
\end{figure}

To generate 100 \textit{AI-generated} texts for our 10~categories in a \textit{basic} way, we prompted ChatGPT ``Generate a text on the following topic: $<topic>$''. To obtain 100~\textit{AI-rephrased} texts for the respective categories in a \textit{basic} way, we used the following prompt: ``Rephrase the following text: $<$text from Wikipedia article$>$'' for every sample. Both prompts are illustrated in Figure~\ref{fig:basicGeneration}. We used the original Wikipedia text excerpts to gather \textit{human-generated} texts. To ensure that we did not accidentally take text generated by ChatGPT, we only used text from Wikipedia articles created before November 2022, when ChatGPT was released to the public. 
The statistics of the \textit{human-generated} texts compared to the \textit{basic} \textit{AI-generated} and \textit{AI-rephrased} texts are summarized in Table~\ref{table:statistics-basic}.

\begin{table}[ht]
\footnotesize
\begin{tabular}{@{}llllllllll@{}}
\toprule
    & \multicolumn{3}{c}{\textbf{Human}}     & \multicolumn{3}{c}{\textbf{AI-generated}}        & \multicolumn{3}{c}{\textbf{AI-rephrased}} \\ 
       \textbf{Category} & \textbf{P} & \textbf{S} & \textbf{W} & \textbf{P} & \textbf{S} & \textbf{W} & \textbf{P} & \textbf{S} & \textbf{W}  \\ \midrule
    Biology & 44& 188& 3739& 54& 139& 2500& 21& 96& 1899\\
    Chemistry & 44& 167&3590 & 56& 140&2684 & 28& 129& 2539\\         
    Geography & 35& 167& 3386& 60& 167& 3006& 27&114& 2540\\
    History & 43& 189& 4578& 61& 148& 3017& 26& 146& 3205\\
    IT & 40& 141&2916 & 51&129 & 2624& 24&91 & 1872\\
    Music &39 & 191& 4177& 53& 154& 2701& 27& 137& 2900\\
    Politics &43 & 172& 4298&56 & 131& 2866& 25& 104& 2341\\
    Religion & 40& 171& 3796& 51& 138& 2684& 25& 108& 2409\\
    Sports & 51& 204& 4692&59 &143 & 2904& 30& 128& 2913\\
    Visual arts &36 & 147& 3165& 54& 136& 2686& 22& 85& 2024\\
    \bottomrule
\end{tabular}
\caption{\textit{Basic AI-Generated/Rephrased} Text \\(P = \#paragraphs, S = \#sentences, W = \#words).}
\label{table:statistics-basic}
\end{table}


\subsection{Advanced AI-Generated Texts}

\begin{figure}[h!]
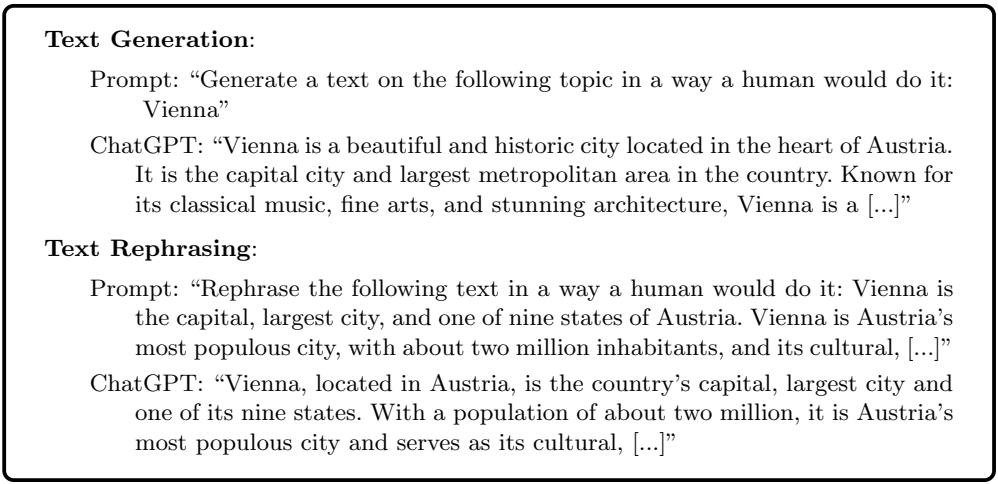

\small
  \begin{tcolorbox}[colback=white,colframe=black]

\textbf{Text Generation}:
\vspace{0.15cm} 

\setlength{\leftskip}{0.6cm}
\hangafter=1
\hangindent=0.7cm
\noindent
     Prompt: ``Generate a text on the following topic in a way a human would do it: Vienna''\vspace{0.1cm} 

\hangafter=1
\hangindent=0.6cm
\noindent
    ChatGPT: ``Vienna is a beautiful and historic city located in the heart of Austria. It is the capital city and largest metropolitan area in the country. Known for its classical music, fine arts, and stunning architecture, Vienna is a [...]''   

\vspace{0.2cm}
\setlength{\leftskip}{0cm}
\textbf{Text Rephrasing}:
\vspace{0.15cm} 

\setlength{\leftskip}{0.6cm}
\hangafter=1
\hangindent=0.6cm
\noindent    
    Prompt:  ``Rephrase the following text in a way a human would do it: Vienna is the capital, largest city, and one of nine states of Austria. Vienna is Austria's most populous city, with about two million inhabitants, and its cultural, [...]''\vspace{0.1cm} 

\hangafter=1
\hangindent=0.6cm
\noindent    
    ChatGPT: ``Vienna, located in Austria, is the country's capital, largest city and one of its nine states. With a population of about two million, it is Austria's most populous city and serves as its cultural, [...]''

  \end{tcolorbox}
  \caption{Prompt and ChatGPT's Response: Advanced Text Generation \& Rephrasing.}
  \label{fig:advancedGeneration}
\end{figure}

Additionally, our intention was to investigate the more \textit{advanced} case where the AI was told to write or rephrase the text in a way that a human would not realize it was generated by an AI, as depicted in Figure~\ref{fig:advancedGeneration}. To get 100 \textit{advanced}~\textit{AI-generated} example texts, we asked ChatGPT ``Generate a text on the following topic in a way a human would do it: $<topic>$'' for the 10 topics of each category. 
Additionally, we collected 100 \textit{advanced}~\textit{AI-rephrased} example texts by asking ChatGPT ``Rephrase the following text in a way a human would do it: $<$text from Wikipedia article$>$'' for the 10 topics of each category. 

The statistics of the \textit{human-generated} texts compared to the \textit{advanced} \textit{AI-generated} and \textit{AI-rephrased} texts are summarized in Table~\ref{table:statistics-advanced}.

\begin{table}[h!]
\footnotesize
\begin{tabular}{@{}llllllllll@{}}
\toprule
    & \multicolumn{3}{c}{\textbf{Human}}     & \multicolumn{3}{c}{\textbf{AI-generated}}        & \multicolumn{3}{c}{\textbf{AI-rephrased}} \\ 
    \textbf{Category} & \textbf{P} & \textbf{S} & \textbf{W} & \textbf{P} & \textbf{S} & \textbf{W} & \textbf{P} & \textbf{S} & \textbf{W}  \\ \midrule
    Biology     & 44 & 188 & 3739 & 47 & 111 & 2057 & 18 & 79  & 1487 \\
    Chemistry   & 44 & 167 & 3590 & 48 & 124 & 2374 & 22 & 103 & 1859 \\
    Geography   & 35 & 167 & 3386 & 49 & 136 & 2575 & 24 & 106 & 2028 \\
    History     & 43 & 189 & 4578 & 52 & 132 & 2583 & 22 & 113 & 2415 \\
    IT          & 40 & 141 & 2916 & 51 & 128 & 2538 & 16 & 86  & 1541 \\
    Music       & 39 & 191 & 4177 & 48 & 128 & 2426 & 21 & 109 & 2145 \\
    Politics    & 43 & 172 & 4298 & 52 & 127 & 2672 & 25 & 111 & 2300 \\
    Religion    & 40 & 171 & 3796 & 48 & 122 & 2623 & 28 & 120 & 2488 \\
    Sports      & 51 & 204 & 4692 & 53 & 143 & 2685 & 31 & 118 & 2407 \\
    Visual arts & 36 & 147 & 3165 & 43 & 119 & 2238 & 23 & 95  & 1931 \\ 
    \bottomrule
\end{tabular}
\caption{\textit{Advanced AI-Generated/Rephrased} Text \\(P = \#paragraphs, S = \#sentences, W = \#words).}
\label{table:statistics-advanced}
\end{table}


\section{Our Features for the Classification of Human- and AI-Generated Texts}\label{sec:features}



For the classification, we implemented the feature categories \textit{perplexity features}, \textit{semantic features}, \textit{list lookup features}, \textit{document features}, \textit{error-based features}, \textit{readability features}, \textit{AI feedback features}, and \textit{text vector features} which performed specifically well in the related work plus new features which have not been analyzed so far. In this section, we will describe each feature category. The features from all categories are summarized in Table \ref{table:featuresummary } at the end of this section.

In ge\subsection{Perplexity-Based Features} 

\textit{Perplexity} is a measure of how well a language model is able to predict a sequence of words~\cite{Vu2010RapidBO}. In other words, it measures how surprised the language model is when it encounters a new sequence of words. A lower perplexity indicates that the language model is better at predicting the next word in a sequence.

When it comes to distinguishing between \textit{human-generated} and \textit{AI-generated} texts, one key difference is that \textit{human-generated} text tends to be more varied and unpredictable than \textit{AI-generated} text. Human writers can use their creativity, knowledge, and experience to produce texts that are full of unexpected word combinations, ideas, and structures. On the other hand, \textit{AI-generated} text is often based on statistical patterns and rules and can be more predictable and repetitive. 
Therefore, researchers like \cite{gehrmann-etal-2019-gltr}, \cite{mitrovic2023chatgpt} and \cite{guo2023close} used perplexity as a feature to distinguish between \textit{human-generated} and \textit{AI-generated} text. 

In our study, we investigated two perplexity features: The mean perplexity (\textit{$PPL_{mean}$}) is calculated by taking the average perplexity across all the sentences in a corpus of text. 
The maximum perplexity (\textit{$PPL_{max}$}) is the highest perplexity that the language model encounters when processing the corpus of text. It represents the most difficult sentence or sequence of words for the language model to predict. In our implementation, we used a Natural Language Toolkit (NLTK)~\cite{bird-loper-2004-nltk} script to compute the perplexities of the texts using a GPT-2 model\footnote{\url{https://github.com/openai/gpt-2}} as \cite{guo2023close} and \cite{mitrovic2023chatgpt}.

\begin{figure}[ht]
  \begin{center}
  \includegraphics[width=1.0\linewidth]{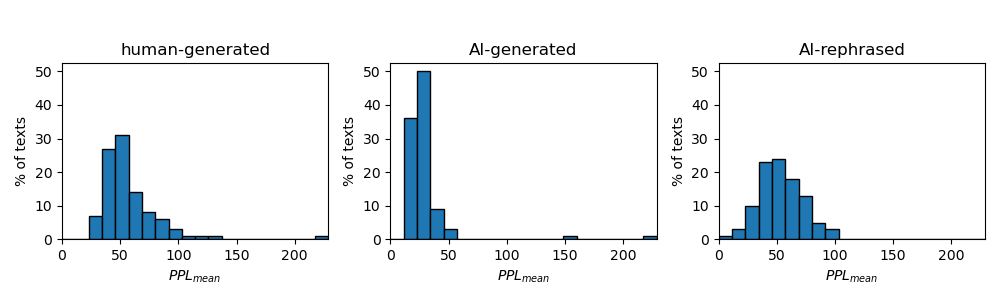}
  \caption{$PPL_{mean}$ Distribution.}
  \label{fig:PPL_mean}
  \end{center}
\end{figure}

Figure~\ref{fig:PPL_mean} compares the \textit{$PPL_{mean}$} distributions in the \textit{human-generated}, \textit{AI-generated} and \textit{AI-rephrased} texts. A high percentage of \textit{AI-generated} texts with values around 25 have lower perplexities than \textit{human-generated} texts with perplexities closer to 50. However, this is not the case for \textit{AI-rephrased} texts, suggesting that this feature will have more problems with \textit{AI-rephrased} than with \textit{AI-generated} texts. 


\subsection{Semantic Features}

\textit{Semantic features} refer to the attributes or properties of words or phrases that can be used to represent the meaning of the words or phrases. These properties can be the sentiment polarity and the degree of subjectivity or objectivity. 

While \cite{mitrovic2023chatgpt} and \cite{guo2023close} used the sentiment polarity as a feature to distinguish between \textit{human-} and \textit{AI-generated} text---to the best of our knowledge ---we are the first to analyze the degree of objectivity and subjectivity as a feature to differ \textit{human-generated} and \textit{AI-generated} texts. 

Sentiment analysis is the process of automatically detecting a sentiment from textual information and presenting the information in classes such as \textit{negative}, \textit{neutral} or \textit{positive}~\cite{Wankhade:2022,Rakhmanov+Schlippe:2022,MabokelaSchlippe:2022b} or with a \textit{sentiment score}. Applying sentiment analysis to a text can help to distinguish if it has been \textit{human-generated} or \textit{AI-generated} as reported in \cite{mitrovic2023chatgpt} and \cite{guo2023close}.

We analyzed two semantic features: First, we applied the sentiment analysis system from TextBlob\footnote{\url{https://textblob.readthedocs.io/en/dev/quickstart.html\#sentiment-analysis}}, a Python library for text processing operations, to retrieve a sentiment polarity score ($sentiment_{polarity}$) between -1 and +1, where -1 represents a very negative text and +1 a very positive text. Second, we used the same Python library to retrieve a subjectivity score ($sentiment_{subjectivity}$) between 0 and +1, where 0 represents a very objective text and +1 a very subjective text.


\begin{figure}[h!]
  \begin{center}
  \includegraphics[width=1.0\linewidth]{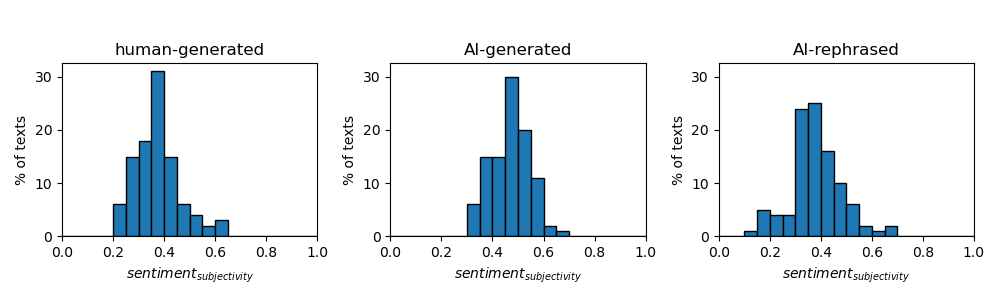}
  \caption{$sentiment_{subjectivity}$ Distribution.}
  \label{fig:sentiment_subjectivity}
  \end{center}
\end{figure}

Figure~\ref{fig:sentiment_subjectivity} illustrates that more \textit{AI-generated} and \textit{AI-rephrased} texts have higher $sentiment_{subjectivity}$ scores than \textit{human-generated} texts. We explain this distribution by the fact that ChatGPT was fine-tuned using reinforcement learning from human feedback~\cite{OpenAI:ChatGPT}.

\subsection{List Lookup Features}

\textit{List lookup features} provide information about the category of a word or character~\cite{nadeau2007survey}. For example, if a word is found in a stop word list (e.g., ``a'', ``an'', ``the'', ``of''), we know that it is a stop word. \cite{Shijaku:2023} and \cite{kumarage2023stylometric} could classify \textit{human-} and \textit{AI-generated} texts using the number of stop words ($stopWord_{count}$) and the number of special characters ($specialChar_{count}$) as features. 

Consequently, we used both features for our classification. As a new promising list lookup feature, we included the number of discourse markers ($discourseMarker_{count}$), such as ``however'', ``furthermore'', or ``moreover''. Additionally, we took the absolute and relative numbers of repetitions of the article's title ($titleRepetition_{count}$, $titleRepetition_{relative}$) since we detected that \textit{AI-generated} text often repeats keywords from the title. 

\begin{figure}[h!]
  \begin{center}
  \includegraphics[width=1.0\linewidth]{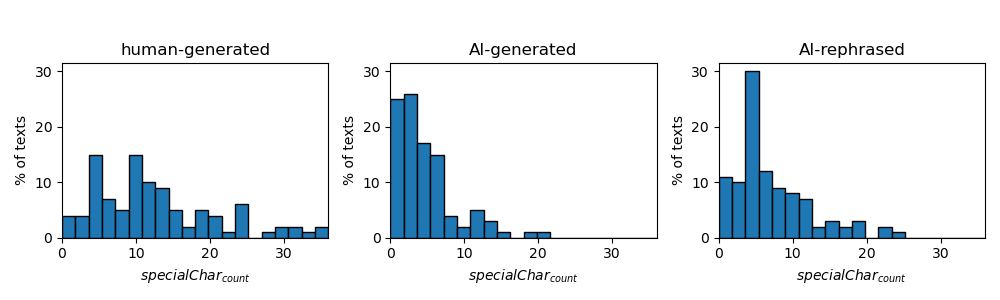}
  \caption{$specialChar_{count}$ Distribution.}
  \label{fig:specialChar_count}
  \end{center}
\end{figure}

Figure~\ref{fig:specialChar_count} visualizes the $specialChar_{count}$ as a representative of the \textit{list lookup features}. The figure indicates that the $specialChar_{count}$ is more widely distributed when the text is \textit{human-generated}.

\subsection{Document Features}

\textit{Document features} are defined by the content and the structure of a document~\cite{nadeau2007survey}. Document features can go beyond single-word and multi-word expressions containing meta-information and corpus statistics (multiple occurrences, local syntax, word frequency, etc.).

In our experiment, we used \textit{document features} related to the frequencies of words, sentences, punctuation marks, characters, and part-of-speech tags which were successful in the classification of \textit{human-} and \textit{AI-generated} texts in~\cite{guo2023close}, \cite{kumarage2023stylometric} and \cite{zaitsu2023distinguishing}. Since in \cite{kumarage2023stylometric} the standard deviation of words and sentences performed well, we also included the standard deviation of the number of unique words per sentence ($uniqWordsPerSentence_{stdev}$) as a new feature. In addition, the number of quotation marks ($quotation_{count}$) is used, as we found that AI produces fewer quotation marks.


\begin{figure}[h!]
  \begin{center}
  \includegraphics[width=1.0\linewidth]{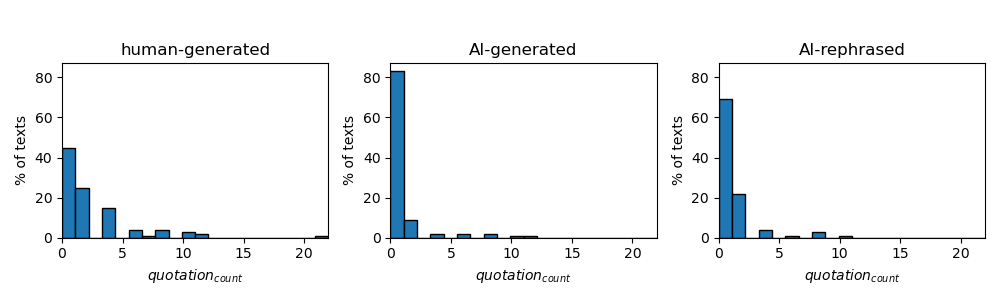}
  \caption{$quotation_{count}$ Distribution.}
  \label{fig:quotation_count}
  \end{center}
\end{figure}


For example, Figure~\ref{fig:quotation_count} illustrates that \textit{AI-generated} and \textit{AI-rephrased} texts contain few to no quotation marks, with over 80\% and over 60\% of texts having no quotation marks, respectively. \textit{Human-generated} text on the other hand has one quotation mark in over 20\% and four quotation marks in over 15\% of our text examples.

\subsection{Error-Based Features}

We observed that in \textit{AI-generated} text fewer spelling and grammar errors occur than in \textit{human-generated} text. Therefore, we introduce \textit{error-based features} as a new feature category and tested the number of spelling and grammar errors ($grammarError_{count}$) as well as the number of multiple blanks ($multiBlank_{count}$) as features from this category. To detect the spelling and grammar errors, we used \textit{LanguageTool}\footnote{\url{https://github.com/jxmorris12/language\_tool\_python}}, an open-source grammar tool, also known as the spellchecker for OpenOffice. We detected multiple blanks using regular expressions. 

\begin{figure}[h!]
  \begin{center}
  \includegraphics[width=1.0\linewidth]{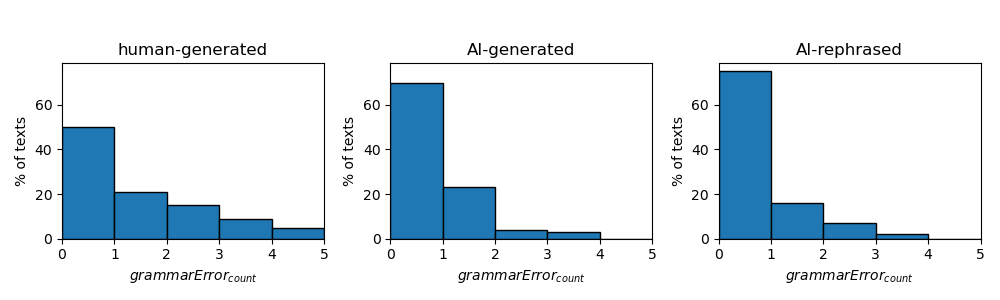}
  \caption{$grammarError_{count}$ Distribution.}
  \label{fig:grammar_error_count}
  \end{center}
\end{figure}

Figure~\ref{fig:grammar_error_count} demonstrates the distribution of $grammarError_{count}$ in the \textit{human-generated}, \textit{AI-generated} and \textit{AI-rephrased} texts. We observe that \textit{LanguageTool} detects more spelling and grammar errors in the \textit{human-generated} than in the \textit{AI-generated} and \textit{AI-rephrased} texts.

\subsection{Readability Features}

Since \textit{readability features} were among the top 5 in \cite{Shijaku:2023}, we also use them for our detection. 
Following \cite{Shijaku:2023}, we implemented the \textit{Flesch Reading Ease} score ($fleschReadingEase$) and \textit{Flesch–Kincaid Grade Level} ($fleschKincaidGradeLevel$). The \textit{Flesch Reading Ease} measures the ease of readability of a text, with higher scores indicating greater ease of reading and lower scores indicating greater difficulty~\cite{Flesch1948ANR}. The \textit{Flesch–Kincaid Grade Level} formula     provides a numerical rating equivalent to a U.S. grade level~\cite{Kincaid1975DerivationON}. This allows educators, guardians, librarians, and others to assess the comprehensibility level of different texts and books with greater ease. The \textit{Flesch Reading Ease} and \textit{Flesch–Kincaid Grade Level} scores are calculated according to a formula that includes the number of words, sentences and syllables~\cite{Flesch1948ANR,Kincaid1975DerivationON}.


\begin{figure}[h!]
  \begin{center}
  \includegraphics[width=1.0\linewidth]{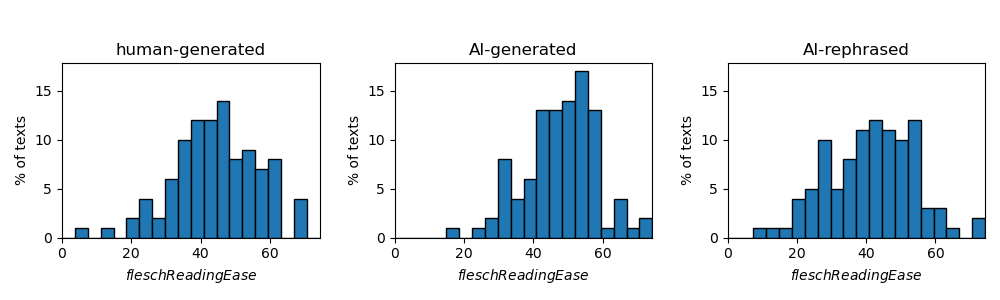}
  \caption{$fleschKincaidGradeLevel$ Distribution.}
  \label{fig:flesch_reading_ease}
  \end{center}
\end{figure}


Figure~\ref{fig:flesch_reading_ease} depicts the \textit{Flesch–Kincaid Grade Level} score distribution in our data set. While a higher number of \textit{AI-generated} texts is on a higher level between 50--60, the $fleschKincaidGradeLevel$ distributions of \textit{human-generated} and \textit{AI-rephrased} text look comparable.

\subsection{AI Feedback Features}

Another novel feature that---to the best of our knowledge---has not yet been used in the detection of \textit{AI-generated} text is the \textit{AI feedback feature}.

\begin{figure}[h!]
  \begin{center}
  \includegraphics[width=1.0\linewidth]{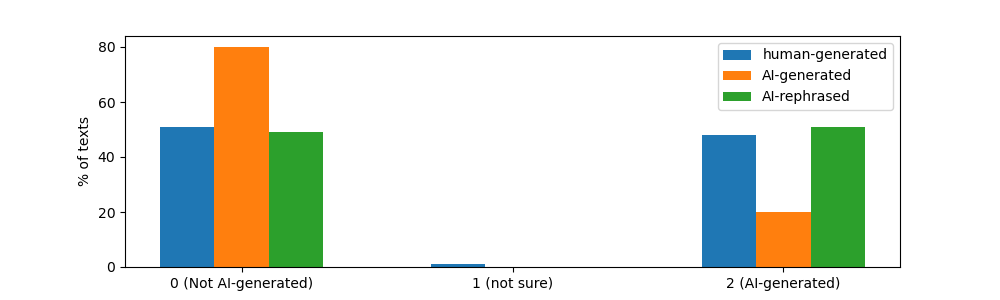}
  \caption{$AIFeedback$ Distribution.}
  \label{fig:ai_feedback}
  \end{center}
\end{figure}

For this feature, we asked ChatGPT directly if it generated a text. If ChatGPT answers 'yes', we assign the value 2 to the feature, if it answers 'no', we assign the value 0. In case ChatGPT answers that it is not sure, we assign 1. However, looking at the distribution of the numerical values, Figure~\ref{fig:ai_feedback} shows that the $AIFeedback$ feature does not seem to discriminate between \textit{AI-generated} and \textit{AI-rephrased} text.


\subsection{Text Vector Features}

To classify the texts using their content, we also analyzed \textit{text vector features}. \textit{TF-IDF} performed well in \cite{Shijaku:2023} and \cite{solaiman2019release}. To take advantage of the benefits of the semantic vector space, we also experimented with \textit{Sentence-BERT}~\cite{reimers-gurevych-2019-sentence}. As we detected that \textit{AI-generated} text often contains repetitive phrases or patterns, we also computed the average distance of Sentence-BERT vectors (\textit{Sentence-BERT-dist}) to detect repetitions as the word embeddings of similar sentences are closer in the semantic vector space.

\subsection{Summary of Our Analyzed Features}

In our experiments, we analyzed a total of 37~features which can be grouped into 8~different categories. Besides those features which have already been studied in related analyses, we included 10~new features. All features which were subject to our analyses are summarized in Table \ref{table:featuresummary }. 

\begin{table}[ht]
\footnotesize
\begin{tabular}{@{}lllr@{}}
\toprule
\textbf{Category} & \textbf{Feature} & \textbf{Description} & \textbf{Reference} \\ \midrule
Perplexity & $PPL_{mean}$ & mean PPL &  \cite{gehrmann-etal-2019-gltr}\cite{mitrovic2023chatgpt}\cite{guo2023close} \\
 & $PPL_{max}$ & maximum PPL & \cite{gehrmann-etal-2019-gltr}\cite{mitrovic2023chatgpt}\cite{guo2023close} \\ \midrule
Semantic & $sentiment_{polarity}$ & degree of positivity/negativity [-1,+1] & \cite{mitrovic2023chatgpt}\cite{guo2023close} \\
 & $sentiment_{subjectivity}$ & degree of subjectivity [0,+1] & new \\ \midrule
List Lookup &$stopWord_{count}$ & number of stop words & \cite{Shijaku:2023} \\
&$specialChar_{count}$ & number of special characters & \cite{kumarage2023stylometric}\\
&$discourseMarker_{count}$ & number of discourse markers & new\\
&$titleRepetition_{count}$ & absolute repetitions of title & new\\
&$titleRepetition_{relative}$ & relative repetitions of title & new\\\midrule
Document  & $wordsPerParagraph_{mean}$ & \diameter number of words per paragraph & \cite{kumarage2023stylometric} \\
&$wordsPerParagraph_{stdev}$ & stdev of $wordsPerParagraph$ & \cite{kumarage2023stylometric}\\ 
&$sentencesPerParagraph_{mean}$ & \diameter number of sentences per paragraph & \cite{kumarage2023stylometric} \\
&$sentencesPerParagraph_{stdev}$ & stdev of $sentencesPerParagraph$ & \cite{kumarage2023stylometric}\\
&$wordsPerSentence_{mean}$ & \diameter number of words per sentence & \cite{kumarage2023stylometric} \\
&$wordsPerSentence_{stdev}$ & stdev of $wordsPerSentence$ & \cite{kumarage2023stylometric}\\
&$uniqWordsPerSentence_{mean}$ & \diameter number of unique words per sentence & \cite{Shijaku:2023} \\
&$uniqWordsPerSentence_{stdev}$ & stdev of $uniqWordsPerSentence$ & new \\
&$words_{count}$ & number of running words & \cite{guo2023close}\cite{Shijaku:2023}\cite{kumarage2023stylometric} \\
&$uniqWords_{count}$ & number of unique words & \cite{kumarage2023stylometric} \\
&$uniqWords_{relative}$ & relative number of unique words & \cite{kumarage2023stylometric} \\
&$paragraph_{count}$ & number of paragraphs & \cite{kumarage2023stylometric} \\
&$sentence_{count}$ & number of sentences & \cite{kumarage2023stylometric} \\
&$punctuation_{count}$ & number of punctuation marks & \cite{kumarage2023stylometric} \\
&$quotation_{count}$ & number of quotation marks & new \\
&$character_{count}$ & number of characters & \cite{kumarage2023stylometric} \\
&$uppercaseWords_{relative}$ & relative number of words in uppercase & \cite{Shijaku:2023}\\
&$personalPronoun_{count}$ & absolute number of personal pronouns & \cite{mitrovic2023chatgpt} \\
&$personalPronoun_{relative}$ & relative number of personal pronouns & \cite{mitrovic2023chatgpt} \\
&$POSPerSentence_{mean}$ & \diameter number of unique POS-tags/sentence & \cite{guo2023close}\cite{kumarage2023stylometric}\cite{zaitsu2023distinguishing} \\
\midrule
 Error-Based& $grammarError_{count}$ & number of spelling/grammar errors & new \\
&$multiBlank_{count}$ & number of multiple blanks & new\\  \midrule
Readability & $fleschReadingEase$ & Flesch Reading Ease score [0-100]  & \cite{Shijaku:2023}\cite{Flesch1948ANR} \\
&$fleschKincaidGradeLevel$ & Readability as U.S. grade level [0-100] & \cite{Shijaku:2023}\cite{Kincaid1975DerivationON}\\ \midrule
AI Feedback & $AIFeedback$ & Ask AI if text was generated by AI & new\\ \midrule
Text Vector &\textit{TF-IDF} & 500-dim TF-IDF vector of 1-/2-grams  & \cite{Shijaku:2023}\cite{solaiman2019release}\\
&\textit{Sentence-BERT} & \diameter Sentence-BERT vector & \cite{reimers-gurevych-2019-sentence}\\
&\textit{Sentence-BERT-dist} & \diameter distance of Sentence-BERT vectors & new\\
\bottomrule
\end{tabular}
\caption{Summary of our Features for the Classification of Generated Texts.}
\label{table:featuresummary }
\end{table}

\section{Experiments and Results}\label{sec:experiments}

In this chapter, we will describe our experiments with the different feature categories and three different classification approaches: The two more traditional approaches XGBoost and random forest (RF) as well as a neural network-based approach with multilayer perceptrons (MLP).

As in other studies like \cite{guo2023close,kumarage2023stylometric,mitrovic2023chatgpt}, we evaluated our systems' classification performance with accuracy (\textit{Acc}) and F1-score (\textit{F1}). Tables~\ref{tab:resultsbasic} and \ref{tab:resultsbasic:AI-rephrased} show the \textit{Acc} and \textit{F1} of detecting the \textit{basic} way of text generation and rephrasing, i.e., without any additional instructions. Tables~\ref{tab:resultsAdvanced:AI-generated} and \ref{tab:resultsAdavanced:AI-rephrased} refer to the \textit{advanced} way, i.e., where the AI was told to write or rephrase the text in a way that a human would not realize it was generated by an AI. 
First, we built \textit{basic text generation detection systems} which were trained, fine-tuned, and tested with our \textit{human-generated} and \textit{basic} \textit{AI-generated} texts. Second, we implemented \textit{basic text rephrasing detection systems} which were trained, fine-tuned, and tested with our \textit{human-generated} and \textit{basic} \textit{AI-rephrased} texts. Third, we built \textit{advanced text generation detection systems} which were trained, fine-tuned, and tested with our \textit{human-generated} and \textit{advanced} \textit{AI-generated} texts. Finally, we built \textit{advanced text rephrasing detection systems} which were trained, fine-tuned, and tested with our \textit{human-generated} and \textit{advanced} \textit{AI-rephrased} texts. 
To provide stable results, we performed a 5-fold cross-validation, randomly dividing our corpus in each fold into 80\% for training, 10\% as a validation set to optimize the hyperparameters, and an unseen test set containing 10\% of the texts. The numerical values in all tables are the average of the test set results. The best performances are highlighted in bold. For a comparison with the state-of-the-art technology, we additionally report GPTZero's performances on our texts.

Table~\ref{tab:resultsbasic} demonstrates that the best performing feature categories are the combination of traditional and our new features from the \textit{document} category ($Document_{traditional+new}$) and the features from the \textit{text vector} category ($Text Vector_{traditional+new}$). With 97\%~\textit{Acc} and 97\%~\textit{F1} $Document_{traditional+new}$ performs substantially better with MLP than XGBoost and RF, while $Text Vector_{traditional+new}$ is most successful with RF (\textit{Acc}=95.0\%, \textit{F1}=94.9\%). Most of our systems were able to outperform GPTZero ($Acc_{GPTZero}$ = 76.0\%, $F1_{GPTZero}$ = 78.9\%). Our best system $All_{traditional+new}$ even performs better than GPTZero by 28.9\% relative in \textit{Acc} and 24.2\% relative in \textit{F1}.

\begin{table}[ht]
\footnotesize
\begin{tabular}{@{}lllllll@{}}
\toprule
& \multicolumn{2}{c}{\textbf{XGBoost}} & \multicolumn{2}{c}{\textbf{RF}} & \multicolumn{2}{c}{\textbf{MLP}}   \\ 
\textbf{Feature Category} & \textbf{Acc} & \textbf{F1} & \textbf{Acc} & \textbf{F1} & \textbf{Acc} & \textbf{F1}\\
\midrule
$Perplexity_{traditional}$& 83.0\% &  82.2\% &  \textbf{87.0\%} &  \textbf{85.3\%} &  82.0\% &  82.1\%\\
\midrule
$Semantic_{traditional}$& 62.0\% &  62.3\% &  66.0\% &  63.6\%&  65.0\% &  61.6\%\\ 
$Semantic_{traditional+new}$& 72.0\% &  72.9\% &  \textbf{75.0\%} &  \textbf{75.6\%}&  73.0\% &  72.3\%\\ \midrule
$List Lookup_{traditional}$& 77.0\% &  78.0\% &  82.0\% &  83.3\%&  \textbf{84.0\%} &  \textbf{83.7\%}\\
$List Lookup_{traditional+new}$& 83.0\% &  82.8\% &  80.0\% &  81.1\%&  81.0\% &  82.9\%\\ \midrule
$Document_{traditional}$& 90.0\% &  90.9\% &  91.0\% &  91.4\%&  94.0\% &  94.1\%\\
$Document_{traditional+new}$& 90.0\% &  90.9\% &  93.0\% &  93.3\%&  \textbf{97.0\%} &  \textbf{97.0\%}\\ \midrule
$ErrorBased_{new}$& 55.0\% &  61.7\% &  55.0\% &  61.7\%&  \textbf{56.0\%} &  \textbf{63.9\%}\\ \midrule 
$Readability_{traditional}$& 60.0\% &  56.3\% &  \textbf{63.0\%} &  \textbf{59.3\%} &  60.0\% &  56.8\%\\ \midrule
$AI Feedback_{new}$& 62.0\% &  67.1\% &  62.0\% &  67.1\%&  \textbf{62.0\%} &  \textbf{68.1\%}\\ \midrule
$Text Vector_{traditional}$& 90.0\% &  89.9\% &  \textbf{95.0\%} &  94.7\%&  86.0\% &  86.3\%\\
$Text Vector_{traditional+new}$& 90.0\% &  89.9\% &  \textbf{95.0\%} &  \textbf{94.9\%}&  81.0\% &  80.6\%\\ \midrule
$All_{traditional}$& 92.0\% &  92.7\% &  97.0\% &  97.0\%&  89.0\% &  89.0\%\\
$All_{traditional+new}$ & 90.0\% &  90.9\% & \textbf{98.0\%} & \textbf{98.0\%}&  87.0\% &  87.8\%\\
\bottomrule
\end{tabular}
\caption{Results for Basic Text Generation: XGBoost vs. RF vs. MLP ($Acc_{GPTZero}$ = 76.0\%, $F1_{GPTZero}$ = 78.9\%).}
\label{tab:resultsbasic}
\end{table}

Table~\ref{tab:resultsbasic:AI-rephrased} indicates that the results of the \textit{basic text rephrasing detection systems} are consistently worse than the results of the \textit{basic text generation detection systems}. The best performing feature categories are $List Lookup_{traditional}$ (\textit{Acc}=73.0\%, \textit{F1}=74.6\%), $Document_{traditional}$ (\textit{Acc}=75.0\%, \textit{F1}=73.9\%), and $Text Vector_{traditional+new}$ (\textit{Acc}=79.0\%, \textit{F1}=78.2\%). 
We observe that for $ErrorBased_{new}$ the \textit{Acc} and \textit{F1} values of our three classifiers are the same. This is due to the fact that $ErrorBased_{new}$ has only 2~dimensions and the classifiers then decide for the same classification. This time XGBoost outperforms RF and MLP. All our systems were able to outperform GPTZero ($Acc_{GPTZero}$=43.0\%, $F1_{GPTZero}$=27.8\%). Our best system $All_{traditional}$ ($Acc$=79.0\%, $F1$=78.9\%) performs much better than GPTZero by 83.7\% relative in \textit{Acc} and even 183.8\% relative in \textit{F1}.

\begin{table}[h!]
\footnotesize
\begin{tabular}{@{}lllllll@{}}
\toprule
& \multicolumn{2}{c}{\textbf{XGBoost}} & \multicolumn{2}{c}{\textbf{RF}} & \multicolumn{2}{c}{\textbf{MLP}}   \\ 
\textbf{Feature Category} & \textbf{Acc} & \textbf{F1} & \textbf{Acc} & \textbf{F1} & \textbf{Acc} & \textbf{F1}\\
\midrule
$Perplexity_{traditional}$& 52.0\% &  48.7\% &  55.0\% &  54.6\%&  \textbf{56.0\%}&  \textbf{63.2\%}\\
\midrule
$Semantic_{traditional}$& 63.0\% &  61.1\% &  66.0\% &  66.0\%&  59.0\% &  61.7\%\\ 
$Semantic_{traditional+new}$& \textbf{66.0\%} &  \textbf{64.4\%} &  66.0\% &  64.3\%&  52.0\% &  54.3\%\\ \midrule
$List Lookup_{traditional}$& 72.0\% &  \textbf{74.6\%} &  69.0\% &  69.5\%&  \textbf{73.0\%} &  74.2\%\\
$List Lookup_{traditional+new}$& 72.0\% &  73.7\% &  66.0\% &  64.9\%&  64.0\% &  63.9\%\\ \midrule
$Document_{traditional}$& \textbf{75.0\%} &  \textbf{73.9\%} &  73.0\% &  73.0\%&  73.0\% &  71.2\%\\
$Document_{traditional+new}$& 72.0\% &  70.9\% &  69.0\% &  68.2\%&  74.0\% &  73.4\%\\ \midrule
$ErrorBased_{new}$& \textbf{62.0\%} &  \textbf{68.0\%} &  \textbf{62.0\%} &  \textbf{68.0\%} &  \textbf{62.0\%} &  \textbf{68.0\%} \\ \midrule
$Readability_{traditional}$& \textbf{54.0\%} &  \textbf{51.1\%} &  \textbf{54.0\%} &  47.8\%&  50.0\% &  50.2\%\\ 
\midrule
$AI Feedback_{new}$& \textbf{52.0\%} &  \textbf{50.9\%} &  50.0\% &  39.8\%&  45.0\% &  30.1\%\\ \midrule
$Text Vector_{traditional}$& 75.0\% &  73.2\% &  77.0\% &  72.2\%&  68.0\% &  63.7\%\\
$Text Vector_{traditional+new}$& \textbf{79.0\%} &  \textbf{78.2\%} &  75.0\% &  71.0\%&  69.0\% &  65.1\%\\ \midrule
$All_{traditional}$& \textbf{79.0\%} &  \textbf{78.9\%} &  73.0\% &  71.6\%&  66.0\% &  65.6\%\\
$All_{traditional+new}$ & 77.0\% &  77.6\% &  71.0\% & 69.8\%&  72.0\% &  71.9\%\\ 
\bottomrule
\end{tabular}
\caption{Results for Basic Text Rephrasing: XGBoost vs. RF vs. MLP ($Acc_{GPTZero}$ = 43.0\%, $F1_{GPTZero}$ = 27.8\%).}
\label{tab:resultsbasic:AI-rephrased}
\end{table}

Table~\ref{tab:resultsAdvanced:AI-generated} shows that the results of our \textit{advanced text generation detection systems} are almost as good as those of our \textit{basic text generation detection systems} which demonstrates that the detection of the \textit{advanced AI-generated} text is not a major challenge for our features. 
The best performing feature categories are $Text Vector_{traditional+new}$ (\textit{Acc}=97.0\%, \textit{F1}=96.9\%), $Document_{traditional}$ (\textit{Acc}=93.0\%, \textit{F1}=93.6\%), $Perplexity_{traditional}$ (\textit{Acc}=85.0\%, \textit{F1}=83.8\%), and $List Lookup_{traditional+new}$ (\textit{Acc}=83.0\%, \textit{F1}=84.8\%).
Again, among XGBoost, RF, and MLP, no classifier shows the best results across all feature categories. Some systems are better than GPTZero ($Acc_{GPTZero}$=79.0\%, $F1_{GPTZero}$=82.7\%). Our best system $Text Vector_{traditional+new}$ (\textit{Acc}=97.0\%, \textit{F1}=96.9\%) outperforms GPTZero by 22.8\% relative in \textit{Acc} and 17.2\% relative in \textit{F1}.

\begin{table}[h!]
\footnotesize
\begin{tabular}{@{}lllllll@{}}
\toprule
& \multicolumn{2}{c}{\textbf{XGBoost}} & \multicolumn{2}{c}{\textbf{RF}} & \multicolumn{2}{c}{\textbf{MLP}}   \\ 
\textbf{Feature Category} & \textbf{Acc} & \textbf{F1} & \textbf{Acc} & \textbf{F1} & \textbf{Acc} & \textbf{F1}\\
\midrule
$Perplexity_{traditional}$& 83.0\% &  82.2\% &  \textbf{85.0\%} &  \textbf{83.8\%}&  83.0\% &  82.6\%\\
\midrule
$Semantic_{traditional}$& 68.0\% &  65.9\% &  68.0\% &  69.0\%&  72.0\% &  70.3\%\\ 
$Semantic_{traditional+new}$& 75.0\% &  71.1\% &  \textbf{76.0\%} &  \textbf{75.1\%}&  73.0\% &  70.2\%\\ \midrule
$List Lookup_{traditional}$& 75.0\% &  76.7\% &  75.0\% &  75.3\%&  78.0\% &  79.0\%\\
$List Lookup_{traditional+new}$& \textbf{83.0\%} &  \textbf{84.8\%} &  82.0\% &  82.6\%&  73.0\% &  73.2\%\\ \midrule
$Document_{traditional}$& 90.0\% &  90.7\% &  \textbf{93.0\%} &  \textbf{93.6\%}&  90.0\% &  89.4\%\\
$Document_{traditional+new}$& 90.0\% &  90.7\% &  91.0\% &  91.8\%&  92.0\% &  91.8\%\\ \midrule
$ErrorBased_{new}$& \textbf{62.0\%} &  \textbf{71.7\%} &  \textbf{62.0\%} &  \textbf{71.7\%}&  59.0\% &  67.8\%\\ \midrule
$Readability_{traditional}$& 60.0\% &  59.7\% &  59.0\% &  56.8\%&  \textbf{65.0\%} &  \textbf{63.2\%}\\ 
\midrule
$AI Feedback_{new}$& \textbf{66.0\%} &  \textbf{71.1\%} &  \textbf{66.0\%} &  \textbf{71.1\%}&  \textbf{66.0\%} &  \textbf{71.1\%}\\ \midrule
$Text Vector_{traditional}$& 90.0\% &  89.1\% &  90.0\% &  89.6\%&  79.0\% &  79.2\%\\
$Text Vector_{traditional+new}$& 90.0\% &  89.1\% &  \textbf{97.0\%} &  \textbf{96.9\%}&  75.0\% &  73.8\%\\ \midrule
$All_{traditional}$& 89.0\% &  90.0\% &  \textbf{95.0\%} &  95.0\%&  86.0\% &  86.0\%\\
$All_{traditional+new}$ & 93.0\% &  94.0\% &  \textbf{95.0\%} &  \textbf{95.9\%}&  84.0\% &  82.5\%\\ 
\bottomrule
\end{tabular}
\caption{Results for Advanced Text Generation: XGBoost vs. RF vs. MLP ($Acc_{GPTZero}$ = 79.0\%, $F1_{GPTZero}$ = 82.7\%).}
\label{tab:resultsAdvanced:AI-generated}
\end{table}

Table~\ref{tab:resultsAdavanced:AI-rephrased} indicates that the results of the \textit{advanced text rephrasing detection systems} are worse than the \textit{advanced text generation detection systems} but even slightly better than the results of the \textit{basic text rephrasing detection systems} which demonstrates that the detection of the \textit{advanced AI-rephrased} text is not a major challenge for our features. The best performing feature categories are $Text Vector_{traditional}$ (\textit{Acc}=80.0\%, \textit{F1}=77.6\%), $List Lookup_{traditional+new}$ (\textit{Acc}=76.0\%, \textit{F1}=75.5\%) and $Document_{traditional+new}$ (\textit{Acc}=77.0\%, \textit{F1}=77.5\%). Among XGBoost, RF and MLP, no classifier shows best results across all feature categories. Again, all our systems are better than GPTZero ($Acc_{GPTZero}$=52.0\%, $F1_{GPTZero}$=45.8\%). Our best system $All_{traditional+new}$ (\textit{Acc}=82.0\%, \textit{F1}=81.7\%) performs much better than GPTZero by 57.7\% relative in \textit{Acc} and even 78.4\% relative in \textit{F1}. 

\begin{table}[h!]
\footnotesize
\begin{tabular}{@{}lllllll@{}}
\toprule
& \multicolumn{2}{c}{\textbf{XGBoost}} & \multicolumn{2}{c}{\textbf{RF}} & \multicolumn{2}{c}{\textbf{MLP}}   \\ 
\textbf{Feature Category} & \textbf{Acc} & \textbf{F1} & \textbf{Acc} & \textbf{F1} & \textbf{Acc} & \textbf{F1}\\
\midrule
$Perplexity_{traditional}$& \textbf{66.0\%} &  65.6\% &  65.0\% &  \textbf{65.8\%}&  60.0\% &  63.7\%\\
\midrule
$Semantic_{traditional}$& 58.0\% &  53.6\% &  61.0\% &  59.4\%&  \textbf{64.0\%} &  \textbf{67.1\%}\\ 
$Semantic_{traditional+new}$& 55.0\% &  56.3\% &  63.0\% &  61.5\%&  61.0\% &  63.6\%\\ \midrule
$List Lookup_{traditional}$& 73.0\% &  73.3\% &  75.0\% &  75.3\%&  69.0\% &  63.3\%\\
$List Lookup_{traditional+new}$& \textbf{76.0\%} &  \textbf{75.5\%} &  75.0\% &  75.3\%&  72.0\% &  70.4\%\\ \midrule
$Document_{traditional}$& \textbf{77.0\%} &  76.7\% &  76.0\% &  77.0\%&  76.0\% &  75.4\%\\
$Document_{traditional+new}$& 76.0\% &  74.9\% &  76.0\% &  76.2\%&  \textbf{77.0\%} &  \textbf{77.5\%}\\ \midrule
$ErrorBased_{new}$& \textbf{62.0\%} &  \textbf{71.7\%} &  \textbf{62.0\%} &  \textbf{71.7\%}&  55.0\% &  62.2\%\\ \midrule
$Readability_{traditional}$& 58.0\% &  55.0\% &  \textbf{67.0\%} &  66.0\%&  \textbf{67.0\%} &  \textbf{68.0\%}\\ 
\midrule
$AI Feedback_{new}$& \textbf{58.0\%} &  \textbf{61.7\%} &  \textbf{58.0\%} &  \textbf{61.7\%}&  \textbf{58.0\%} &  \textbf{61.7\%}\\ \midrule
$Text Vector_{traditional}$& 77.0\% &  72.6\% &  \textbf{80.0\%} &  \textbf{77.6\%}&  61.0\% &  56.0\%\\
$Text Vector_{traditional+new}$& 71.0\% &  66.5\% &  78.0\% &  75.0\%&  73.0\% &  71.3\%\\ \midrule
$All_{traditional}$& 81.0\% &  80.6\% &  76.0\% &  76.2\%&  71.0\% &  69.0\%\\
$All_{traditional+new}$ & \textbf{82.0\%} &  \textbf{81.7\%} &  76.0\% &  76.3\%&  77.0\% &  75.2\%\\ 
\bottomrule
\end{tabular}
\caption{Results for Advanced Text Rephrasing: XGBoost vs. RF vs. MLP ($Acc_{GPTZero}$ = 52.0\%, $F1_{GPTZero}$ = 45.8\%).}
\label{tab:resultsAdavanced:AI-rephrased}
\end{table}

\section{Conclusion and Future Work}\label{sec:conclusion}

In this paper, we explored traditional and new features to detect \textit{AI-generated} texts. We produced a new data corpus covering 10 school topics. We were able to achieve an F1-score of 98.0\% for \textit{basic human-generated}/\textit{AI-generated} texts and an F1-score of 78.9\% for \textit{basic human-generated}/\textit{AI-rephrased} texts. Furthermore, we reported an F1-score of 96.9\% for \textit{advanced human-generated}/\textit{AI-generated} texts and an F1-score of 81.7\% for \textit{advanced human-generated}/\textit{AI-rephrased} texts. Our best \textit{basic} text rephrasing detection system even outperforms GPTZero by 183.8\% relative in F1-score. Our results show that the new features can help to improve classification performance. As tools like ChatGPT are nowadays easy to access, generated exams or student papers have become a serious issue as this can undermine students' learning outcomes and academic integrity. Our results can make an important contribution to the detection of \textit{AI-generated} texts and to help teachers to identify generated texts. 

So far we investigated features to detect \textit{AI-generated} and \textit{AI-rephrased} text in English. Additionally, we plan to classify text in other languages. Moreover, we demonstrated that the performances of the systems which combine all features are very close. Consequently, in future work, it is interesting to consider a system combination that has the potential to even further increase performance. 
While we have already analyzed two types of prompts---the \textit{basic} and the \textit{advanced} variants, our goal is to investigate further variants and their impact on classification performance.


\bibliography{sn-bibliography}

\end{document}